\documentclass{article}
\usepackage{spconf,amsmath,epsfig,amssymb}

\title{GENERALIZED COMPRESSION DICTIONARY DISTANCE \\
AS UNIVERSAL SIMILARITY MEASURE}
\twoauthors
{Andrey Bogomolov
\sthanks{Thanks to Telecom Italia for funding.}
}
{
SKIL Telecom Italia Lab, University of Trento, \\
Via Sommarive, 5, I-38123, Trento, Italy \\
E-mail: andrey.bogomolov@unitn.it
}
{Bruno Lepri, Fabio Pianesi
\sthanks{Performed the work while at Fondazione Bruno Kessler}
}
{Fondazione Bruno Kessler \\
Via Sommarive, 18, I-38123, Trento, Italy \\
E-mail: \{lepri, pianesi\}@fbk.eu
}

\begin{document}
%\ninept
%
\maketitle
\begin{abstract}
We present a new similarity measure based on information theoretic
measures which is superior than Normalized Compression Distance for
clustering problems and inherits the useful properties of conditional Kolmogorov
complexity.

We show that Normalized Compression Dictionary Size and Normalized Compression
Dictionary Entropy are computationally more efficient, as the need to perform
the compression itself is eliminated. Also they scale linearly with exponential
vector size growth and are content independent.

We show that normalized compression dictionary distance is compressor
independent, if limited to lossless compressors, which gives space for
optimizations and implementation speed improvement for real-time and big data
applications.

The introduced measure is applicable for machine learning tasks of
parameter-free unsupervised clustering, supervised learning such as
classification and regression, feature selection, and is applicable for big data
problems with order of magnitude speed increase. 
\end{abstract}

\section{Introduction}
\label{sec:Introduction}

The similarity measure between objects is fundamental for machine learning
tasks. Most similarity measures require prior assumptions on the statistical
model and/or the parameters limits.

For most applications in computational social science, economics, finance, human
dynamics analysis this implies a certain risk of being biased or not to account
for partially observed source signal fundamental properties
\cite{taleb2012antifragile}.

For more technical applications such as digital signal processing,
telecommunications and remote sensing, given that the signal could be observed
and modelled, we face the problems of noise, features representation and
algorithmic efficiency.

We easily agree with the following outcome of 20 trials of fair coin 
toss ``01111011001101110001'', but we do not accept the result
``00000000000000000000''. However, both results have equal chances given that the
fair coin model assumption holds. This is a common example of paradoxes in
probability theory, but our reaction is caused by the belief that the first
sequence is complicated, but the second is simple \cite{Muchnik1998263}.
A second example of human-inspired limitations is ``Green Lumber Fallacy''
introduced by Nassim Nicholas Taleb. It is a kind of fallacy that a person
"mistaking the source of important or even necessary knowledge, for another less
visible from the outside, less tractable one''. 
Mathematically, it could be expressed as we use an incorrect function which, by
some chance, returns the correct output, such that $g(x)$ is mixed with
$f(x)$. The root of the fallacy is that ``although people may be focusing on the
right things, due to complexity of the thing, are not good enough to figure it out intellectually''
\cite{taleb2012antifragile}.

Despite the psychological limitations and fallacies by which we, human, reason
and develop the models, during the last few decades there were developed a
number of robust methods to enhance model generalization properties and resistance to noise, 
such as filtering, cross validation, boosting, bootstrapping, bagging,
random forests \cite{hastie01statisticallearning}.
The most promising approach to the challenging paradigm of
approaching antifragility uses Kolmogorov complexity theory
\cite{Kolmogorov1998387} and concepts of Computational Irreducibility or the Principle of Computational Equivalence, introduced by Steven Wolfram
\cite{DBLP:books/daglib/0006578}.
Unfortunately Kolmogorov complexity is uncomputable for a general case. For
practical applications we have to implement the algorithms which run on
computers having Turing machine properties.

\section{Methodology}
\label{sec:Methodology}

For defining similarity measure which uses Kolmogorov complexity researchers
introduced Information Distance measure, which is defined as a distance between
strings $x$ and $y$ as the length of the shortest program $p$ that computes $x$ from $y$ and vice versa.
The Information Distance is absolute and to obtain a similarity metric
Normalized Information Distance ($NID$) was introduced:

\begin{equation}
NID(x,y) = 
\frac 
 {max\{K(x|y), K(y|x)\}} 
 {max\{K(x), K(y)\}}  
\end{equation}

Unfortunately, Normalized Information Distance is also uncomputable for a
general case, as it is dependent on uncomputable Kolmogorov complexity measure
\cite{1412045}. For approximating NID in a practical environment -- Normalized
Compression Distance ($NCD$) was developed, based on based on a real-world
lossless abstract compressor $C$ \cite{1362909}:

\begin{equation}
NCD(x,y) = 
\frac 
 {C(xy) - min\{C(x), C(y)\}} 
 {max\{C(x), C(y)\}}  
\end{equation}

Daniele Cerra and Mihai Datcu introduced another approximation metric -- Fast
Compression Distance ($FCD$), which is applicable to medium-to-large datasets:

\begin{equation}
FCD(x,y) = 
\frac 
 {\lvert D(x) \rvert - \cap(D(x), D(y))} 
 {\lvert D(x) \rvert},  
\end{equation}

where $\lvert D(x) \rvert$ and $\lvert D(y) \rvert$ are the sizes of the
relative dictionaries, represented by the number of entries they contain, 
and $\cap(D(x), D(y))$ is the number of patterns which are found in both
dictionaries. FCD accounts for the number of patterns which are found in both
dictionaries extracted during compression by Lempel-Ziv-Welch algorithm,
and reduces the computational effort by computing the intersection between
dictionaries, which represents the joint compression step performed in NCD
\cite{Cerra:2012:FCS:2109696.2109997}.

We found that the number of patterns which exist in the both dictionaries is
dependent on the compression algorithm used. Also the dictionary of $x$ and
$y$ set intersection could be coded with different symbols, as the frequencies
of strings could be different in $x$, $y$ and $\cap(x, y)$, which leads to less
accurate approximation. The size of a compression dictionary does not account
for the symbol frequency properties of the dictionary and the size of possible
algorithmic ``description of the string in some fixed universal description language'',
which is the essence of Kolmogorov complexity. That means that we loose a lot of
information about $x$ and $y$ if we compute only the size of a compression
dictionary.

In order to overcome this problem we introduce Generalized Compression
Dictionary Distance (GCDD) metric, which is defined as:

\begin{equation}
\mathbf{GCDD}(x,y) = 
\frac 
 {\mathbf{\Phi} (x \cdot y) - min\{\mathbf{\Phi}(x), \mathbf{\Phi}(y)\}} 
 {max\{\mathbf{\Phi}(x), \mathbf{\Phi}(y)\}},  
\end{equation}

where $\mathbf{\Phi} (x \cdot y)$ is functional characteristics of the
compression dictionary extracted from concatination of $x$ and $y$ byte arrays.
GCDD returns an n-dimensional vector, which characterizes the
conditional similarity measure between  $x$ and $y$. Each dimension of GCDD
represents a real valued function.

The algorithmic complexity of the proposed solution is proportional to:

\begin{equation}
O_{GCDD(x,y)} \rightarrow 
	k  m_x \log m_y,
\end{equation}

where $m_x$ and $m_y$ is the dictionary size of $x$ and $y$,
and $k$ is the constant dependent on the dimensionality of the resulting vector.  

In comparison, the algorithmic complexity of the other measures are:

\begin{equation}
O_{FCD(x,y)} \rightarrow 
	m_x \log m_y,
\end{equation}
\begin{equation}
O_{NCD(x,y)} \rightarrow 
	(n_x + n_y) \log (m_x + m_y),
\end{equation}

which shows asymptotically small increase in computational time for GCDD but
preserving informational gain through additional $x$ and $y$ characteristics
transfer.

\section{Experimental Results and Discussion}
\label{sec:ExperimentalResults}

The implementation of Generalized Compression Dictionary Distance prototype was
done in The Java Platform, Standard Edition 8 for double precision 64-bit input
vectors and working with the Huffman Coding and Lempel-Ziv-Welch compression for
byte array case applying the approach of binary-to-string encoding.

The experiments were run on ``Synthetic Control Chart Time Series'' -- 
a well known dataset published in UCI Machine Learning
Repository\cite{Asuncion+Newman:2007}, which contains control charts
synthetically generated by Alcock and Manolopoulos \cite{Alcock99time-seriessimilarity} 
for the six different classes of time series: normal, cyclic, increasing trend,
decreasing trend, upward shift and downward shift.

Experimental results show that Normalized Compression
Dictionary Size and Normalized Compression Dictionary Entropy, as examples of
GCDD, give more stable and accurate results for time-series clustering problem
when tested on heterogeneous input vectors, than NCD and other 
traditional distance (e.i. dissimilarity) measures, such as
euclidean distance.

\begin{figure}[ht] \centering{
\includegraphics[scale=0.44]{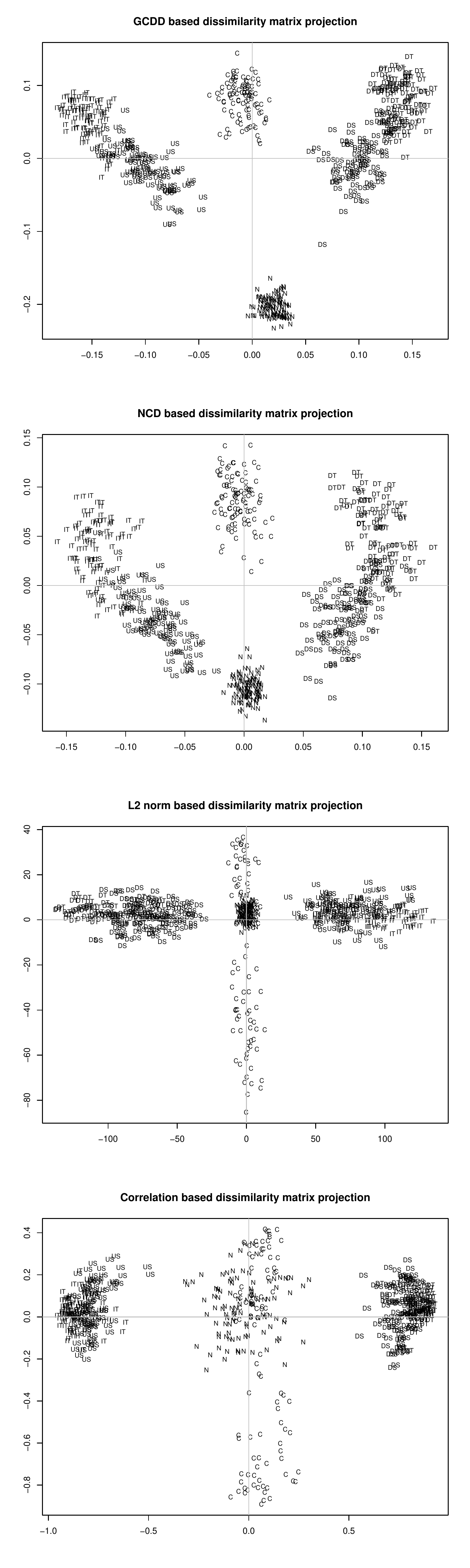}}
\caption{Fast Methods Comparison}
\label{fig:dissimilarity}
\end{figure}

Experimental results shown in the Fig.\ref{fig:dissimilarity} are produced from
abovementioned collection of $I$ time series vectors, on which $4$ distance
function are defined (GCDD, NCD, L2-norm, Pearson correlation).

Applying a distance function for each pairwise vectors, the dissimilarity matrix
is consructed, such that:
\begin{equation}
\Delta := 
\begin{pmatrix}
\delta_{1,1} & \delta_{1,2} & \cdots & \delta_{1,I} \\
\delta_{2,1} & \delta_{2,2} & \cdots & \delta_{2,I} \\
\vdots 		 & \vdots 		& \vdots \\
\delta_{I,1} & \delta_{I,2} & \cdots & \delta_{I,I}
\end{pmatrix}.
\end{equation}

Then multidimensional scaling is performed. Given dissimilarity matrix 
$\Delta$, we find $I$ vectors $x_1,\ldots,x_I \in \mathbb{R}^N$ such that
$\|x_i - x_j\| \approx \delta_{i,j}$ for all $i,j\in {1,\dots,I}$, where
$\|\cdot\|$ is a vector norm.

On the plots we use the following symbols to encode vectors of time series
trend types: N - normal, C - cyclic, IT - increasing trend, DT - decreasing
trend, US - upward shift and DS - downward shift.

From the Fig.\ref{fig:dissimilarity} we see, that GCDD based distance
metric efficiently groups time series on a hyperplane thus increasing separation
ability.

It has similar properties as NCD, and much better than L2-norm and Pearson
correlation based, where the time series vectors are mixed.

Then we run the experiments with computationally intensive state-of-the-art
methods for time series clustering, such as: 
(1) autocorrelation based method,
(2) Linear Predictive Coding based as proposed by Kalpakis, 2001
\cite{Kalpakis2001}, 
(3) Adaptive Dissimilarity Index based \cite{Chouakria} and 
(4) ARIMA based (Piccolo, 1990) \cite{piccolo1990distance}.

\begin{figure}[ht] \centering{
\includegraphics[scale=0.44]{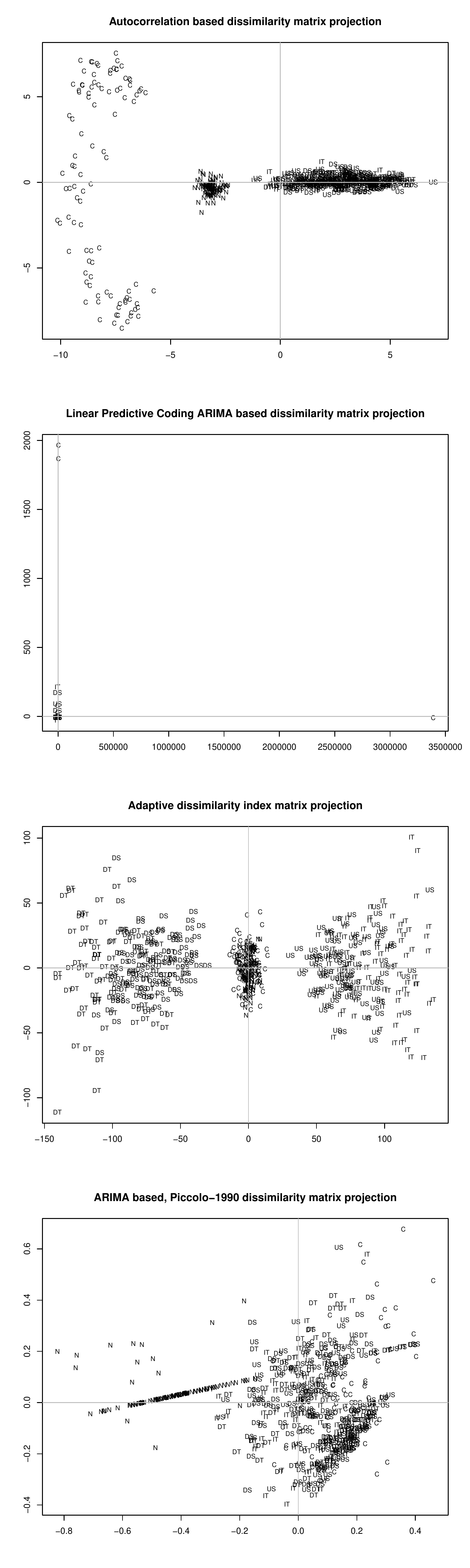}}
\caption{Numerically Intensive Methods Comparison}
\label{fig:dissimilarity2}
\end{figure}

From the Fig.\ref{fig:dissimilarity2} we see, that numerically intensive
methods do not enhance much the separation ability, which is computed applying
GCDD based distance metric.
Also these distance methods require much more computation time and are not
applicable for big data problems.

Further more, the result proposed in this paper could be used for unsupervised
clustering, supervised classification and feature representation for deep learning tasks 
given the nice properties of GCDD, such as (1) it scales linearly with
exponential growth of the input vector size and (2) it is content independent, as the semantics is coded
inside the extracted dictionary itself.

The future research steps include testing the concept on diverse data sets,
including image processing data and using the GCDD output for different machine
learning tasks.

\bibliographystyle{IEEEbib}
\bibliography{strings,refs}

\end{document}